\documentclass{article}
\pdfoutput=1 

\usepackage[nonatbib,preprint]{neurips_2020}

\usepackage[utf8]{inputenc} 
\usepackage[T1]{fontenc}    
\usepackage[pdfstartview=FitH,
            CJKbookmarks=true,
            bookmarksnumbered=true,
            bookmarksopen=true,
            colorlinks,
            linkcolor=red,
            anchorcolor=blue,
            citecolor=gray]{hyperref}
\usepackage{url}            
\usepackage{booktabs}       
\usepackage{amsfonts}       
\usepackage{nicefrac}       
\usepackage{microtype}
\usepackage[round,sort,comma,authoryear]{natbib}     

\usepackage{booktabs}
\usepackage{ulem}
\usepackage{makecell,multirow,diagbox}  
\usepackage{float}
\usepackage{adjustbox}
\usepackage{lscape}
\usepackage{amsfonts,amssymb}
\usepackage{amsthm}
\usepackage{amsmath}
\usepackage{amssymb}
\usepackage{color}
\usepackage{xcolor}
\usepackage{algorithm} 
\usepackage{algorithmic}
\usepackage{subfigure}
\usepackage{enumitem}
\definecolor{c_red}{HTML}{BF1616} 
\definecolor{c_green}{HTML}{097609}

\AtBeginDocument{%
  \providecommand\BibTeX{{%
    \normalfont B\kern-0.5em{\scshape i\kern-0.25em b}\kern-0.8em\TeX}}}

\title{Historical Inertia: A Neglected but Powerful Baseline for Long Sequence Time-series Forecasting }

\author{%
  Yue Cui\\
  University of Electronic Science\\ and Technology of China\\
  Chengdu, 611731\\
  \texttt{cuiyue@uestc.edu.cn} \\
  \And
  Jiandong Xie\\
  Cloud BU\\
  Huawei Technologies Co. Ltd. \\
  Chengdu, 611731\\
  \texttt{xiejiandong@huawei.com} \\
  \And
  Kai Zheng\\
  University of Electronic Science\\ and Technology of China\\
  Chengdu, 611731 \\
    \texttt{zhengkai@uestc.edu.cn} \\}

\begin{document}

\maketitle

\begin{abstract}
Long sequence time-series forecasting (LSTF) has become increasingly popular for its wide range of applications. Though superior models have been proposed to enhance the prediction effectiveness and efficiency, it is reckless to neglect or underestimate one of the most natural and basic temporal properties of time-series. In this paper, we introduce a new baseline for LSTF, the historical inertia (HI), which refers to the most recent historical data-points in the input time series. We experimentally evaluate the power of historical inertia on four public real-word datasets. The results demonstrate that up to 82\% relative improvement over state-of-the-art works can be achieved even by adopting HI directly as output. 

\end{abstract}

\section{Introduction}
Time series forecasting, i.e., given historical values of time series and making prediction for future time-slots, can be deemed as one of the main enablers of modern society. An accurate prediction model can benefit a wide-range of applications, e.g., predicting stock prices \citep{TEG,arima}, monitoring traffic flows and electricity consumption \citep{informer,MTGNN,TPA_LSTM,LSTNet}. 

Rather than the typical setting of predicting values of limited number of time-steps, i.e. 48 steps or fewer \citep{MTGNN,TEG,TPA_LSTM,LSTNet}, an emerging line of work focuses on the problem of long sequence time-series forecasting (LSTF), where up to 720 steps can be predicted at a time \citep{informer}. Such an increasing sequence length can be troublesome to most existing works, which are designed for relatively short prediction horizon. 

To deal with the challenges of effectively modeling temporal correlations in long sequence and efficiently operating on long inputs and outputs, the state-of-the-art (SOTA) work Informer \citep{informer} proposes a novel variant of Transformer \citep{attention} to reduce time and space complexity while maintaining prediction accuracy, which is indeed a breakthrough. Despite that the extensive experiments on five real-world datasets demonstrates Informer's superiority to its baselines, the enhanced performance can be limited when considering the baseline of taking the most recent values in inputs as outputs, which can be referred to as the historical inertia (HI). 

In this paper, we first address this issue by providing an experimental evaluation of the proposed baseline HI and SOTA models and on a variety of public real-world datasets, and then make a comprehensive discussion on why HI is powerful and how we can benefit from HI.

\section{Problem and The Proposed Baseline}

\textbf{Long Sequence Time-series Forecasting:} At time $t$, given a $L_x$-length time series as input, i.e., $\mathcal{X}(t)=\{X_1(t),...,X_{L_x}(t)\}$, where $X_i(t) =[x_{i,1}(t),..,x_{i,d_{x}}(t)] \in \mathbb{R}^{d_{x}}, i\in[1,...,{L_x}]$, is the observed univariate ($d_{x}=1$) or multivariate ($d_{x}>1$) variable at the $i$-th time-stamp, the goal of long sequence time-series forecasting (LSTF) is to predict the corresponds $L_y$-length sequence $\Delta$ steps ahead, i.e., $\mathcal{Y}(t)=[Y_{1}(t),..,Y_{L_y}(t)]$, where $Y_i(t) =[y_{i,1}(t),..,y_{i,d_{y}}(t)] \in \mathbb{R}^{d_{y}}$ and $d_{x}\geq d_{y}\geq1$. When $d_{x} = d_{y}$,  $\mathcal{Y}(t)= [X_{L_x+\Delta+1}(t),..,X_{L_x+\Delta+L_y}$ $(t)]$.

\textbf{Historical Inertia:} The historical inertia (HI) baseline takes $L_y$-length subsequence of $\mathcal{X}(t)$ as prediction results, i.e., $\hat{\mathcal{Y}(t)}= [X_{L_x-L_y+1}(t),..,X_{L_x}(t)]$. 

 \begin{figure}[t]
\centering
\includegraphics[width=0.6\linewidth]{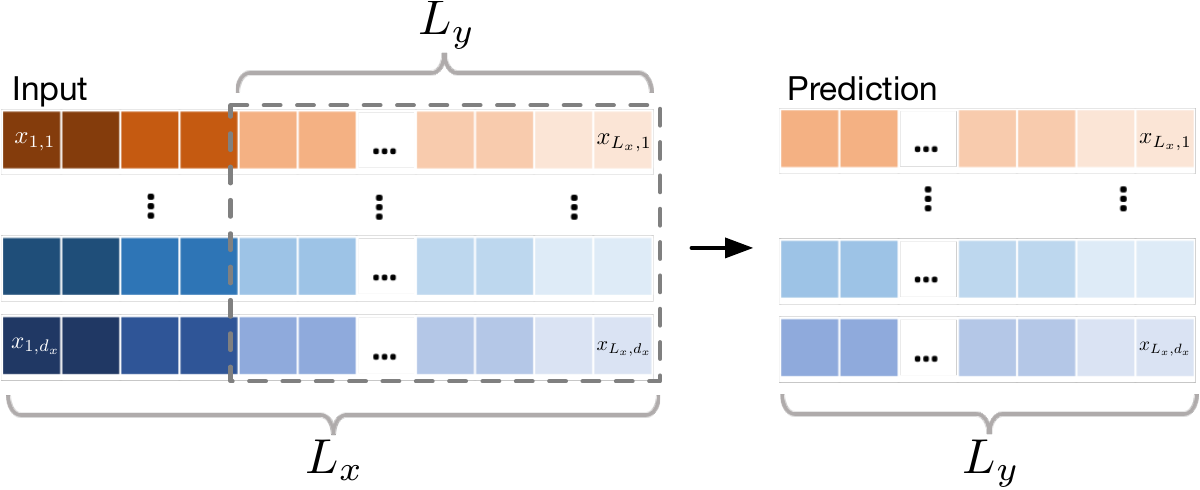}
\caption{The proposed baseline HI, illustrated in the scenario of multivariate time series forecasting. HI directly takes the most recent time steps in the input as prediction. $L_x$ is the input length and $L_y$ is the prediction length. $d_x$ denotes the number of variables in inputs.} 
\label{fig:hi}
\end{figure} 

Note that the HI requires prediction length to be no longer than the input length, i.e., $L_x\geq L_y$, which is not necessary for learning-based LSTF models. Considering that in real application scenarios, the dataset is usually orders of magnitude larger than $L_y$, this condition can be easily achieved. An illustration of the proposed baseline is shown in Figure \ref{fig:hi}.

\section{Experiment and Results}

\subsection{Datasets and Metrics}
We compare HI with SOTA models on four real-world public datasets.

\textbf{ETT} (Electricity Transformer Temperature) \footnote{https://github.com/zhouhaoyi/}: The ETT dataset from the Informer paper contains 2-years electric power deployment collected from two Chinese counties. There are 7 features in total. Three sub-datasets are included in our experiments, i.e. ETTh1 and ETTh2 with an 1-hour sampling frequency and ETThm1 with a 15-min sampling frequency. In the univariate forecasting task, the feature "oil temperature" is chosen as prediction target.

\textbf{Electricity} \footnote{https://github.com/laiguokun/multivariate-time-series-data}: The raw dataset of Electricity is from the UCI Machine Learning Repository \footnote{https://archive.ics.uci.edu/ml/datasets/ElectricityLoadDiagrams20112014}, which contains electricity consumption of 370 clients every 15 minutes from 2011 to 2014. We use the pre-processed dataset from \cite{LSTNet}, which reflects hourly consumption of 321 clients from 2012 to 2014. The last client (column) is used as prediction target in the univariate forecasting task.

Statistics of above datasets can be found in Table \ref{tb:statistics}.  

\setlength{\tabcolsep}{1mm}{
\begin{table}[h]
\caption{Statistics of dataset.}
\centering
\begin{tabular}{llllll}

\toprule
Dataset& \# samples&\# variables &Sample rate\\
\midrule
ETTh1 & 17420 & 7 & 1 hour\\
ETTh2 & 17420 &7 & 1 hour\\
ETTm1 & 69680 &7 & 15 minutes \\
Electricity & 26304 & 321 &1 hour \\
\bottomrule
\end{tabular}
\label{tb:statistics}
\vspace{-0.25cm}
\end{table}}

As a common practice, we evaluate the models by two metrics: Mean Square Error (MSE) and Mean Absolute Error (MAE), which are computed as:

\begin{equation}
MSE=\frac{1}{T}\sum\limits_{t=1}^{T}\frac{1}{ L_y \times d_y}\sum\limits_{i=t_1}^{t_{L_y}}\sum\limits_{j=1}^{d_y}(\hat{y}_{i,j}(t)-y_{i,j}(t))^2,
\end{equation}
\begin{equation}
MAE=\frac{1}{T}\sum\limits_{t=1}^{T}\frac{1}{ L_y \times d_y} \sum\limits_{i=t_1}^{t_{L_y}}\sum\limits_{j=1}^{d_y}|\hat{y}_{i,j}(t)-y_{i,j}(t)|
\end{equation}

where $\hat{y}$ is the prediction output, $y$ is the ground-truth value, $t\in[t_1,t_T]$ is the time instance in test set.

All above settings are consistent with the Informer paper. Note that we eliminate the dataset Weather that is also used in the paper, since only raw data is available and the preprocessing operations are unclear.
\subsection{Competitors}
\subsubsection{Univariate LSTF SOTA Models}
Eight models ranging from traditional statistical methods to recent-proposed deep models are included as competitors for the task of univariate time series forecasting.
\begin{itemize}
\item \textbf{Prophet} \citep{prophet}: A regression model that models common feature of time series in scale-aware way.
\item \textbf{ARIMA} \citep{arima}: An autoregressive integrated moving average-based model for stock price prediction. 
\item \textbf{DeepAR} \citep{DeepAR}: An autoregressive recurrent neural network. 
\item \textbf{LSTMa} \citep{lstma}: A recurrent neural network-based neural machine translation model designed for long sentences.
\item \textbf{Reformer} \citep{reformer}: An efficient variant of Transformer using locality-sensitive hashing and reversible residual layers.
\item \textbf{LogTrans} \citep{logtrans}: An efficient variant of Transformer using convolutional attention and sparse attention.
\item \textbf{Informer} \citep{informer}: An efficient variant of Transformer using ProbSparse self-attention and self-attention distilling.
\item \textbf{Informer-} \citep{informer}: A variant of Informer removing the ProbSparse self-attention mechanism.

\end{itemize}

\subsubsection{Multivariate LSTF SOTA Models}

Besides above mentioned \textbf{LSTMa}, \textbf{Reformer}, \textbf{LogTrans}, \textbf{Informer} and \textbf{Informer-}, 
\begin{itemize}

\item \textbf{LSTNet} \citep{LSTNet}: A deep neural network that combines convolutional neural networks and recurrent neural networks,

\end{itemize}
is used as a competitor in the task of multivariate time series forecasting.
\subsection{Implementation Details}
Basically, we follow the common practice in the community as described in \cite{informer}. $\Delta$ is fixed as 1. Prediction length is set as [24, 48, 168, 366, 720] for ETTh1 and ETTh2, [24, 48, 96, 288, 672] for ETTm1 and [48, 168, 366, 720, 960] for Electricity. We split the ETT datasets into 12:4:4 and Electricity dataset into 15:3:4 for training, validation and test. Above implementation settings are consistent with the Informer paper. Since the method of HI doesn't require training, when the dataset split is fixed, the performance is fixed. Thus, only one iteration is sufficient to compute the final results.
\subsection{Main Results}

Table \ref{tb:mainresults-uni} and Table \ref{tb:mainresults-multi} provide the main experimental results of HI and SOTA models. The best results are highlighted in bold. The last line in each Table calculates HI's relative improvement over the best SOTA model, which is calculated as $(best\_SOTA\_Model-HI)/best\_SOTA\_Model$. Numbers in \textcolor{c_green}{green} indicate positive and in \textcolor{c_red}{red} indicate negative. All reported results are on test set. Besides the results of HI, numbers are referenced from the updated results on the paper of Informer \cite{informer}. We also follow the same scaling strategy as Informer does.

We observe that HI achieves state-of-the-art results in many cases, especially for the task of multivariate forecasting, in which the relative improvement can be up to 82\%. In the following, we discuss experimental results of univariate and multivariate LSTF respectively.
\subsubsection{Univariate LSTF Results}
Table \ref{tb:mainresults-uni} shows that in the task of predicting a single variable over time, HI outperforms SOTA models significantly on ETTh1 and ETTm1 dataset. Informer and its variant almost dominate the ETTh2 dataset while DeepAR, Informer and HI claims part of the best results on the Electricity dataset. The relative improvement brought by HI can be up to 80\% on MSE and 58\% on MAE.

\setlength{\tabcolsep}{0.45mm}{
\begin{table*}[]
\centering
\caption{Summary of univariate long sequence time-series forecasting comparison results.}
\label{tb:mainresults-uni}
\fontsize{6.5}{7.5}\selectfont

\begin{tabular}{cc|ccccc|ccccc|ccccc|ccccc}
\toprule
\multicolumn{2}{c}{Dataset} & \multicolumn{5}{|c}{ETTh1}  & \multicolumn{5}{|c}{ETTh2} & \multicolumn{5}{|c}{ETTm1}&\multicolumn{5}{|c}{Electricity}\\
\midrule 
Method &Metric&24&48&168&336&720&24&48&168&336&720&24&48&96&288&672&48&168&336&720&960\\
\midrule 
\multirow{2}*{Prophet} &MSE& 0.115& 0.168& 1.224& 1.549& 2.735& 0.199& 0.304& 2.145& 2.096& 3.355& 0.120& 0.133&0.194& 0.452& 2.747&0.524& 2.725& 2.246& 4.243& 6.901\\
                        &MAE& 0.275& 0.330& 0.763& 1.820& 3.253& 0.381& 0.462& 1.068& 2.543& 4.664& 0.290& 0.305&0.396& 0.574& 1.174&0.595& 1.273& 3.077& 1.415& 4.264\\

\midrule 
\multirow{2}*{ARIMA}  &MSE& 0.108& 0.175& 0.396& 0.468& 0.659& 3.554& 3.190& 2.800& 2.753& 2.878& 0.090& 0.179&0.272& 0.462& 0.639&0.879& 1.032& 1.136& 1.251& 1.370\\
                        &MAE& 0.284& 0.424& 0.504& 0.593& 0.766& 0.445& 0.474& 0.595& 0.738& 1.044& 0.206& 0.306&0.399& 0.558& 0.697&0.764& 0.833& 0.876& 0.933& 0.982\\
\midrule 
\multirow{2}*{DeepAR}        &MSE&0.107& 0.162& 0.239&0.445& 0.658& 0.098& 0.163& 0.255& 0.604& 0.429& 0.091& 0.219& 0.364& 0.948& 2.437& \textbf{0.204}&\textbf{0.315}& \textbf{0.414}& 0.563& 0.657\\
                            &MAE&0.280& 0.327& 0.422& 0.552& 0.707& 0.263& 0.341&0.414& 0.607& 0.580& 0.243& 0.362& 0.496& 0.795& 1.352&\textbf{0.357}&0.436& 0.519& 0.595& 0.683\\
\midrule 
\multirow{2}*{LSTMa}     &MSE& 0.114&0.193& 0.236& 0.590& 0.683&0.155& 0.190& 0.385& 0.558& 0.640& 0.121& 0.305& 0.287& 0.524& 1.064& 0.493&0.723& 1.212& 1.511& 1.545\\
                            &MAE& 0.272& 0.358& 0.392& 0.698& 0.768& 0.307& 0.348& 0.514& 0.606& 0.681& 0.233& 0.411& 0.420& 0.584& 0.873& 0.539&0.655& 0.898& 0.966& 1.006\\
\midrule 
\multirow{2}*{Reformer}  &MSE& 0.222& 0.284& 1.522& 1.860& 2.112& 0.263& 0.458& 1.029& 1.668& 2.030& 0.095& 0.249& 0.920& 1.108& 1.793& 0.971&1.671& 3.528& 4.891& 7.019\\
                            &MAE& 0.389& 0.445& 1.191& 0.124& 1.436& 0.437& 0.545& 0.879& 1.228& 1.721& 0.228& 0.390& 0.767& 1.245& 1.528& 0.884&1.587& 2.196& 4.047& 5.105\\

\midrule 
\multirow{2}*{LogTrans}       &MSE& 0.103& 0.167&   0.207& 0.230& 0.273& 0.102& 0.169& 0.246& 0.267& 0.303& 0.065& 0.078& 0.199& 0.411& 0.598& 0.280&0.454& 0.514& 0.558& 0.624\\
                        &MAE& 0.259& 0.328& 0.375& 0.398& 0.463& 0.255& 0.348& 0.422& 0.437& 0.493& 0.202& 0.220& 0.386& 0.572& 0.702& 0.429&0.529& 0.563& 0.609& 0.645\\
\midrule
\multirow{2}*{Informer-}  &MSE& 0.092& 0.161& 0.187& 0.215& 0.257& 0.099& 0.159& 0.235& \textbf{0.258}& 0.285& 0.034& 0.066&0.187& 0.409& 0.519& 0.238&0.442& 0.501& 0.543& 0.594\\
                        &MAE& 0.246& 0.322& 0.355& 0.369& 0.421& 0.241& 0.317& 0.390& 0.423& 0.442& 0.160& 0.194&0.384& 0.548& 0.665& 0.368&0.514& 0.552& 0.578& 0.638\\
\midrule 
                          
\multirow{2}*{Informer}   &MSE& 0.098& 0.158& 0.183& 0.222& 0.269& \textbf{0.093}& 0.155& \textbf{0.232}& 0.263& \textbf{0.277}& 0.030& 0.069&0.194& 0.401& 0.512& 0.239&0.447& 0.489& \textbf{0.540}& \textbf{0.582}\\
                        &MAE& 0.247& 0.319& 0.346& 0.387& 0.435& 0.240& 0.314& \textbf{0.389}& \textbf{0.417}& \textbf{0.431}& 0.137& 0.203&0.372& 0.554& 0.644& 0.359&0.503& 0.528& \textbf{0.571}& \textbf{0.608}\\
\midrule
\midrule
\multirow{2}*{HI}   &MSE&\textbf{0.046}  &\textbf{0.069}  &\textbf{0.116}  &\textbf{0.137}  &\textbf{0.186}  &0.095 &\textbf{0.150}  &0.257  &0.318  &0.449  &\textbf{0.023}  &\textbf{0.039}  &\textbf{0.046 }  &\textbf{0.081 }  &\textbf{0.115 } & 0.872 &0.328 &0.415 &1.178 &1.302\\
                    &MAE&\textbf{0.166}  &\textbf{0.210}  &\textbf{0.271}  &\textbf{0.306}  &\textbf{0.351}  &\textbf{0.231}  &\textbf{0.300}  &0.409  &0.465  & 0.549 & \textbf{0.115} & \textbf{0.156} &\textbf{0.167} &\textbf{0.229 } & \textbf{0.270 }&0.690 &\textbf{0.393} &\textbf{0.463} &0.836 &0.894\\
\midrule
\midrule 
\multirow{2}*{Improve}       &MSE& \textcolor{c_green}{50 \%} & \textcolor{c_green}{56\%} &\textcolor{c_green}{37\%} &\textcolor{c_green}{36\%} &\textcolor{c_green}{28\%} &\textcolor{c_red}{2\%} &\textcolor{c_green}{3\%} &\textcolor{c_red}{11\%} &\textcolor{c_red}{23\%} &\textcolor{c_red}{62\%} &\textcolor{c_green}{23\%} &\textcolor{c_green}{41\%} &\textcolor{c_green}{75\%} &\textcolor{c_green}{80\%} &\textcolor{c_green}{78\%} &\textcolor{c_red}{327\%} &\textcolor{c_red}{4\%} &\textcolor{c_red}{0\%} &\textcolor{c_red}{118\%} &\textcolor{c_red}{124\%} \\

                        &MAE&\textcolor{c_green}{33\%} &\textcolor{c_green}{34\%} &\textcolor{c_green}{37\%} &\textcolor{c_green}{17\%} &\textcolor{c_green}{17\%} &\textcolor{c_green}{4\%} &\textcolor{c_green}{4\%} &\textcolor{c_red}{5\%} &\textcolor{c_red}{12\%} &\textcolor{c_red}{27\%} &\textcolor{c_green}{16\%} &\textcolor{c_green}{20\%} &\textcolor{c_green}{55\%} &\textcolor{c_green}{58\%} &\textcolor{c_green}{58\%} &\textcolor{c_red}{93\%} &\textcolor{c_green}{10\%} &\textcolor{c_green}{11\%} &\textcolor{c_red}{46\%} &\textcolor{c_red}{47\%} \\
\bottomrule

\end{tabular}
  \end{table*}

\setlength{\tabcolsep}{0.5mm}{
\begin{table*}[h!]
\centering
\caption{Summary of multivariate long sequence time-series forecasting comparison results.}
\label{tb:mainresults-multi}
\fontsize{6.5}{7.5}\selectfont

\begin{tabular}{cc|ccccc|ccccc|ccccc|ccccc}
\toprule
\multicolumn{2}{c}{Dataset} & \multicolumn{5}{|c}{ETTh1}  & \multicolumn{5}{|c}{ETTh2} & \multicolumn{5}{|c}{ETTm1}&\multicolumn{5}{|c}{Electricity}\\ 
\midrule 
Method &Metric&24&48&168&336&720&24&48&168&336&720&24&48&96&288&672&48&168&336&720&960\\ 
\midrule 
\multirow{2}*{LSTMa} &MSE& 0.650& 0.702& 1.212& 1.424& 1.960& 1.143& 1.671& 4.117& 3.434& 3.963& 0.621& 1.392&1.339& 1.740& 2.736&0.486& 0.574& 0.886& 1.676& 1.591\\ 
                        &MAE& 0.624& 0.675& 0.867& 0.994& 1.322& 0.813& 0.221& 1.674& 1.549& 1.788& 0.629& 0.939&0.913& 1.124& 1.555&0.572& 0.602& 0.795& 1.095& 1.128\\

\midrule 
\multirow{2}*{Reformer}  &MSE& 0.991& 1.313& 1.824& 2.117& 2.415& 1.531& 1.871& 4.660& 4.028& 5.381& 0.724& 1.098&1.433& 1.820& 2.187&1.404& 1.515& 1.601& 2.009& 2.141\\ 
                        &MAE& 0.754& 0.906& 1.138& 1.280& 1.520& 1.613& 1.735& 1.846& 1.688& 2.015& 0.607& 0.777&0.945& 1.094& 1.232&0.999& 1.069& 1.104& 1.170& 1.387\\ 
\midrule 
\multirow{2}*{LogTrans}        &MSE&0.686& 0.766& 1.002&1.362& 1.397& 0.828& 1.806& 4.070& 3.875& 3.913& 0.419& 0.507& 0.768& 1.462& 1.669& 0.355&0.368& 0.373& 0.409& 0.477\\ 
                            &MAE&0.604& 0.757& 0.846& 0.952& 1.291& 0.750& 1.034&1.681& 1.763& 1.552& 0.412& 0.583& 0.792& 1.320& 1.461& 0.418&0.432& 0.439& 0.454& 0.589\\ 
\midrule 
\multirow{2}*{LSTNet}     &MSE& 1.293&1.456& 1.997& 2.655& 2.143&2.742& 3.567& 3.242& 2.544& 4.625& 1.968& 1.999& 2.762& 1.257& 1.917& 0.369&0.394& 0.419& 0.556& 0.605\\ 
                            &MAE& 0.901& 0.960& 1.214& 1.369& 1.380& 1.457& 1.687& 2.513& 2.591& 3.709& 1.170& 1.215& 1.542& 2.076& 2.941& 0.445&0.476& 0.477& 0.565& 0.599\\ 
\midrule 
\multirow{2}*{Informer-}  &MSE& 0.620& 0.692& 0.947& 1.094& 1.241& 0.753& 1.461& 3.485& 2.626& 3.548& \textbf{0.306}& \textbf{0.465}& 0.681& 1.162& 1.231& 0.334&0.353& 0.381& 0.391& 0.492\\ 
                            &MAE& 0.577& 0.671& 0.797& 0.813& 0.917& 0.727& 1.077& 1.612& 1.285& 1.495& \textbf{0.371}& \textbf{0.470}& 0.612& 0.879& 1.103&0.399& 0.420& 0.439& \textbf{0.438}& 0.550\\ 

\midrule 
\multirow{2}*{Informer}       &MSE& 0.577& 0.685&   0.931& 1.128& 1.215& 0.720& 1.457& 3.489& 2.723& 3.467& 0.323& 0.494& 0.678& 1.056& 1.192& 0.344&0.368& 0.381& \textbf{0.406}& \textbf{0.460}\\ 
                        &MAE& 0.549& 0.625&   0.752& 0.873& 0.896& 0.665& 1.001& 1.515& 1.340& 1.473& 0.369& 0.503& 0.614& 0.786& 0.926& 0.393&0.424& 0.431& 0.443& 0.548\\ 
\midrule
\multirow{2}*{HI}   &MSE&\textbf{0.426} &\textbf{0.498}   &\textbf{0.653}   &\textbf{0.690} &\textbf{0.714}   &\textbf{0.266}  &\textbf{0.379} &\textbf{0.572} &\textbf{0.567}  &\textbf{0.635} &1.395   &1.668   &\textbf{0.423}   &\textbf{0.526}  &\textbf{0.655}   & \textbf{0.328} &\textbf{0.212} &\textbf{0.247} &0.469 &0.518\\ 
                    &MAE&\textbf{0.390}  &\textbf{0.423}  &\textbf{0.509}  &\textbf{0.527}  &\textbf{0.563}  &\textbf{0.304}  &\textbf{0.374}   &\textbf{0.481}   &\textbf{0.500}  & \textbf{0.530}  & 0.720 &0.821  &\textbf{0.387}   &\textbf{0.444 }  & \textbf{0.508 }&\textbf{0.329} &\textbf{0.279} &\textbf{0.312} &0.439 &\textbf{0.471}\\ 
\midrule
\midrule 
\multirow{2}*{Improve}       &MSE& \textcolor{c_green}{26\%} &\textcolor{c_green}{27\%} &\textcolor{c_green}{30\%} &\textcolor{c_green}{37\%} &\textcolor{c_green}{41\%} &\textcolor{c_green}{63\%} &\textcolor{c_green}{74\%} &\textcolor{c_green}{82\%} &\textcolor{c_green}{78\%} &\textcolor{c_green}{82\%} &\textcolor{c_red}{356\%} &\textcolor{c_red}{259\%} &\textcolor{c_green}{38\%} &\textcolor{c_green}{50\%} &\textcolor{c_green}{45\%} &\textcolor{c_green}{2\%} &\textcolor{c_green}{40\%} &\textcolor{c_green}{34\%} &\textcolor{c_red}{20\%} &\textcolor{c_red}{13\%}\\

                        &MAE&\textcolor{c_green}{29\%} &\textcolor{c_green}{32\%} &\textcolor{c_green}{32\%} &\textcolor{c_green}{39\%} &\textcolor{c_green}{37\%} &\textcolor{c_green}{54\%} &\textcolor{c_green}{63\%} &\textcolor{c_green}{68\%} &\textcolor{c_green}{61\%} &\textcolor{c_green}{64\%} &\textcolor{c_red}{95\%} &\textcolor{c_red}{75\%} &\textcolor{c_green}{37\%} &\textcolor{c_green}{44\%} &\textcolor{c_green}{45\%} &\textcolor{c_green}{16\%} &\textcolor{c_green}{34\%} &\textcolor{c_green}{28\%} &\textcolor{c_red}{0\%} &\textcolor{c_green}{14\%} \\
\bottomrule

\end{tabular}
\end{table*}}

\subsubsection{Multivariate LSTF Results}
Table \ref{tb:mainresults-multi} compares HI against SOTA models for the task of predicting multiple variables over time. We observe that almost all the best results are achieved by HI and the improvement is significant. In the task of predicting 168 and 720 steps ahead on ETTh2 dataset, competitors' best MSE are 3.242 and 3.467, HI reduces them to 0.572 and 0.635, bringing in up to 82\% relative improvement. 

\subsection{Study of HI}
While Table \ref{tb:mainresults-uni} and Table \ref{tb:mainresults-multi} already demonstrate HI's performance against SOTAs, in this section, we emphasize the potential of HI serving as an effective trick by showing how it can help to improve performance of a basic model. We combine HI with another simple method: multi-layer perceptron (MLP) to further explore the effect of HI. The implementation of MLP is the same for all tests in this section. We use a 2-layer MLP with embedding dimension 200. An 1d batch normalization layer, a ReLU layer and a dropout layer with dropout rate 0.05 are added to each hidden layer of the MLP. We set the batch size as 32. The training epoch is set as 30 with early stopping patience 3 on validation loss, which is defined as MSE. Learning rate is initialized as 0.0003 and will be reduced by half every epoch. For each test we run 5 iterations and report the mean values as the final results, as shown in Table \ref{tb:studyhi-uni} and Table \ref{tb:studyhi-multi}. Informer is also included for comparison.

The very first observation is that MLP itself is a also a strong baseline, which outperforms HI and state-of-the-art models across almost all datasets and prediction lengths. Regardless of this point, we take MLP as a basic model, and evaluate the the ensemble of MLP and HI. We operate weighted summation over MLP's and HI's outputs to get the final prediction. The weights of two models are set as 0.5/0.5. From Table \ref{tb:studyhi-uni} and Table \ref{tb:studyhi-multi}, it could be concluded that this hybrid model can obtain better results in many cases, which is especially evidential for the task of univariate forecasting. MLP + HI brings up to 32\% relative improvement over HI and 45\% relative improvement over MLP on MSE, and 20\%,  27\% relative improvement on MAE.

\setlength{\tabcolsep}{0.5mm}{
\begin{table}
\centering
\caption{Summary of univariate long sequence time-series forecasting comparison results with MLP.}
\label{tb:studyhi-uni}
\fontsize{7}{8}\selectfont
\begin{tabular}{@{}l|l|ll|ll|ll|ll@{}}
\toprule
\multicolumn{2}{l|}{Method}           & \multicolumn{2}{l|}{Informer} & \multicolumn{2}{l|}{HI} & \multicolumn{2}{l|}{MLP}        & \multicolumn{2}{l}{MLP + HI}    \\ \midrule
Dataset                      & Metric & MSE           & MAE           & MSE        & MAE        & MSE            & MAE            & MSE            & MAE            \\ \midrule
\multirow{5}{*}{ETTh1}       & 24     & 0.098         & 0.247         & 0.046      & 0.166      & 0.046          & 0.165          & \textbf{0.037} & \textbf{0.146} \\
                             & 48     & 0.158         & 0.319         & 0.069      & 0.210      & \textbf{0.064} & \textbf{0.193} & 0.104          & 0.265          \\
                             & 168    & 0.183         & 0.346         & 0.116      & 0.271      & \textbf{0.099} & \textbf{0.243} & 0.103          & 0.248          \\
                             & 336    & 0.222         & 0.387         & 0.137      & 0.306      & 0.170          & 0.335          & \textbf{0.093} & \textbf{0.243} \\
                             & 720    & 0.269         & 0.435         & \textbf{0.186}      & \textbf{0.351}      & 0.313          & 0.483          & 0.307 & 0.482 \\ \midrule
\multirow{5}{*}{ETTh2}       & 24     & 0.093         & 0.240         & 0.095      & 0.231      & 0.078          & 0.214          & \textbf{0.074} & \textbf{0.207} \\
                             & 48     & 0.155         & 0.314         & 0.150      & 0.300      & 0.105          & 0.252          & \textbf{0.104} & \textbf{0.249} \\
                             & 168    & 0.232         & 0.389         & 0.257      & 0.409      & 0.185          & 0.337          & \textbf{0.164} & \textbf{0.316} \\
                             & 336    & 0.263         & 0.417         & 0.318      & 0.465      & 0.216          & 0.371          & \textbf{0.194} & \textbf{0.351} \\
                             & 720    & 0.277         & 0.431         & 0.449      & 0.549      & \textbf{0.281} & \textbf{0.428} & 0.314          & 0.451          \\ \midrule
\multirow{5}{*}{ETTm1}       & 24     & 0.030         & 0.137         & 0.023      & 0.115      & 0.020          & 0.110          & \textbf{0.016} & \textbf{0.094} \\
                             & 48     & 0.069         & 0.203         & 0.039      & 0.156      & \textbf{0.029} & \textbf{0.128} & 0.030          & 0.132          \\
                             & 96     & 0.194         & 0.372         & 0.046      & 0.167      & 0.070          & 0.210          & \textbf{0.069} & \textbf{0.208} \\
                             & 288    & 0.401         & 0.554         & \textbf{0.081}      & \textbf{0.229}      & 0.091 & 0.238 & 0.107          & 0.264          \\
                             & 672    & 0.512         & 0.644         & 0.115      & 0.270      & 0.199          & 0.372          & \textbf{0.088} & \textbf{0.227} \\ \midrule
\multirow{5}{*}{Electricity} & 48     & 0.239         & 0.359         & 0.872      & 0.690      & 0.266          & 0.370          & \textbf{0.251} & \textbf{0.354} \\
                             & 168    & 0.447         & 0.503         & 0.328      & 0.393      & 0.275          & 0.372          & \textbf{0.248} & \textbf{0.347} \\
                             & 336    & 0.489         & 0.528         & 0.415      & 0.463      & 0.331          & 0.414          & \textbf{0.300} & \textbf{0.382} \\
                             & 720    & 0.540         & 0.571         & 1.178      & 0.836      & 0.390          & 0.454          & \textbf{0.363} & \textbf{0.453} \\
                             & 960    & 0.582         & 0.608         & 1.302      & 0.894      & \textbf{0.442} & \textbf{0.499} & 0.464          & 0.523          \\ \bottomrule
\end{tabular}

\vspace{-0.25cm}
\end{table}}

\setlength{\tabcolsep}{0.5mm}{
\begin{table}
\centering
\caption{Summary of multivariate long sequence time-series forecasting comparison results with MLP.}
\label{tb:studyhi-multi}
\fontsize{7}{8}\selectfont
\begin{tabular}{@{}l|l|ll|ll|ll|ll@{}}
\toprule
\multicolumn{2}{l|}{Method}           & \multicolumn{2}{l|}{Informer} & \multicolumn{2}{l|}{HI} & \multicolumn{2}{l|}{MLP}        & \multicolumn{2}{l}{MLP + HI}      \\ \midrule
Dataset                      & Metric & MSE           & MAE           & MSE        & MAE        & MSE            & MAE            & MSE            & MAE            \\ \midrule
\multirow{5}{*}{ETTh1}       & 24     & 0.577         & 0.549         & 0.426      & 0.390      & 0.312          & 0.361          & \textbf{0.305} & \textbf{0.353} \\
                             & 48     & 0.685         & 0.625         & 0.498      & 0.423      & 0.353          & 0.386          & \textbf{0.346} & \textbf{0.377} \\
                             & 168    & 0.931         & 0.752         & 0.653      & 0.509      & \textbf{0.450} & 0.451          & 0.453          & \textbf{0.444} \\
                             & 336    & 1.128         & 0.873         & 0.690      & 0.527      & \textbf{0.489} & \textbf{0.484}          & 0.526          & 0.497          \\
                             & 720    & 1.215         & 0.896         & 0.714      & 0.563      & \textbf{0.533} & \textbf{0.526} & 0.581          & 0.552          \\ \midrule
\multirow{5}{*}{ETTh2}       & 24     & 0.720         & 0.665         & 0.266      & 0.304      & 0.186          & 0.279          & \textbf{0.184} & \textbf{0.277} \\
                             & 48     & 1.457         & 1.001         & 0.379      & 0.374      & \textbf{0.247} & \textbf{0.319} & 0.253          & 0.321          \\
                             & 168    & 3.489         & 1.515         & 0.572      & 0.481      & \textbf{0.370} & 0.411          & 0.378          & \textbf{0.406} \\
                             & 336    & 2.723         & 1.340         & 0.567      & 0.500      & \textbf{0.410} & \textbf{0.443} & 0.471          & 0.479          \\
                             & 720    & 3.467         & 1.473         & \textbf{0.635}      & \textbf{0.530}      & 0.797          & 0.648          & 0.698 & 0.595 \\ \midrule
\multirow{5}{*}{ETTm1}       & 24     & 0.323         & 0.369         & 1.395      & 0.720      & 0.227          & 0.298          & \textbf{0.225} & \textbf{0.298} \\
                             & 48     & 0.494         & 0.503         & 1.668      & 0.821      & \textbf{0.298} & \textbf{0.345} & 0.300          & 0.350          \\
                             & 96     & 0.678         & 0.614         & 0.423      & 0.387      & 0.335          & 0.372          & \textbf{0.331} & \textbf{0.369} \\
                             & 288    & 1.056         & 0.786         & 0.526      & 0.444      & \textbf{0.360} & 0.391          & 0.361          & \textbf{0.389} \\
                             & 672    & 1.192         & 0.926         & 0.655      & 0.508      & \textbf{0.438} & \textbf{0.437} & 0.447          & 0.444          \\ \midrule
\multirow{5}{*}{Electricity} & 48     & 0.344         & 0.393         & 0.328      & 0.329      & 0.183          & 0.272          & \textbf{0.178} & \textbf{0.265} \\
                             & 168    & 0.368         & 0.424         & 0.212      & 0.279      & 0.173          & 0.275          & \textbf{0.162} & \textbf{0.258} \\
                             & 336    & 0.381         & 0.431         & 0.247      & 0.312      & 0.186          & 0.291          & \textbf{0.176} & \textbf{0.277} \\
                             & 720    & 0.406         & 0.443         & 0.469      & 0.439      & \textbf{0.219} & \textbf{0.321} & 0.223          & 0.322          \\
                             & 960    & 0.460         & 0.548         & 0.518      & 0.471      & \textbf{0.235} & \textbf{0.335} & 0.243          & 0.339          \\ \bottomrule
\end{tabular}

\vspace{-0.35cm}
\end{table}}

\section{Discussion}
Given above results, it could be concluded that though naive, HI is a strong baseline but unfortunately neglected for comparison in LSTF research. However, it is more important why it is powerful and how we could benefit from it. 

\subsection{Why Historical Inertia Works}
A common belief is that predictable time series should have tractable patterns in phase and magnitude. We credit the very first reason HI is powerful to that it guarantees the outputs are in similar magnitude of the inputs. This is especially true in the scenario of long sequence time-series forecasting, because the temporal patterns of a time-series can be more steady if viewed in the long run.

However, phase is much more tricky. On the one hand, longer time series provide more evident periodic patterns that can not be reflected in short horizons. This increases the chance that the HI be of similar phase as the prediction target, especially in the case that the prediction length is an exact integer multiple of the time series' period when their is any. On the other hand, HI could also badly hurt the prediction results when 1) there is no periodic pattern; 2) the periodic pattern is not included in historical data; 3) or the historical data is in opposite phase as the prediction target. The multivariate prediction results on ETTm1 dataset serves a good evidence of above statements. Since the data was sampled by 15-minute, a prediction length of 24 or 48 is too short to reflect periodic patterns. Therefore, HI performs much worse than SOTA models. However, for predicting length of 96, 288 and 672, where the 1-day (4$\times$24 data-points) period is well covered, the relative improvement surges.
\subsection{Benefit from Historical Inertia}
Being of so much power, HI has the potential to serve as an effective trick. We now discuss possible ways of implementations from the perspectives of post-process and pre-process.
\subsubsection{Hybrid Model}
A model may benefit from combining the basic model's and HI's results in a post-process fashion. For example, the simplest implementation would be making weighted summation of the two prediction sequences as the proposed MLP + HI does.

\subsubsection{AutoML}
The modeling capacity of complex architectures is definitely valuable, but just in some cases the answer to the question can be so simple that might not be answered well when it is complicated by the model. It is desirable that a model's structure or complexity can be adaptable to the input, which is also known as automated machine learning (AutoML) (e.g. \citealt{automl}). A simple implementation could be when a specific dataset is given, the model may first analyze its temporal pattens in a pre-processed way, and then score whether the basic model, HI or some median variants should be used for prediction. 
\section{Conclusion}
In this paper we propose a baseline for LSTF, named HI. It directly takes the most recent time steps in the input as output. Extensive experiments in four public real-world datasets validate the strength of HI across different prediction lengths. We hope HI could serve as a basement and spark future LSTF research.
\normalem
\bibliographystyle{apalike}
\bibliography{refs}

\end{document}